\documentclass[letterpaper, 10 pt, conference]{ieeeconf}  

\usepackage{times}
\usepackage{epsfig}
\usepackage{graphicx}
\usepackage{amsmath}
\usepackage{amssymb}

\usepackage{multirow}
\usepackage{siunitx}
\usepackage{booktabs}
\usepackage{pdfpages}
\usepackage{subcaption}
\usepackage{cite}
\usepackage{arydshln}
\IEEEoverridecommandlockouts                              

\usepackage[T1]{fontenc}

\title{\LARGE \bf
DVLO4D: Deep Visual-Lidar Odometry with Sparse Spatial-temporal Fusion

 \author{Mengmeng Liu$^{1^*}$, Michael Ying Yang$^{2}$, Jiuming Liu$^{3}$, Yunpeng Zhang$^{4}$, Jiangtao Li$^{4}$, \\ Sander Oude Elberink$^{1}$, George Vosselman$^{1}$, Hao Cheng$^{1^*}$
 \thanks{$^{1}$ University of Twente, The Netherlands} 
 \thanks{$^{2}$ University of Bath, UK}
  \thanks{$^{3}$ Shanghai Jiao Tong University, China}
    \thanks{$^{4}$ PhiGent Robotics, China}
    \thanks{$^*$ Corresponding authors: \{m.liu-1; h.cheng-2\}.utwente.nl}
}}

\begin{document}

\maketitle
\thispagestyle{empty}
\pagestyle{empty}

\begin{abstract}

Visual-LiDAR odometry is a critical component for autonomous system localization, yet achieving high accuracy and strong robustness remains a challenge. Traditional approaches commonly struggle with sensor misalignment, fail to fully leverage temporal information, and require extensive manual tuning to handle diverse sensor configurations. To address these problems, we introduce DVLO4D, a novel visual-LiDAR odometry framework that leverages sparse spatial-temporal fusion to enhance accuracy and robustness.
Our approach proposes three key innovations: (1) Sparse Query Fusion, which utilizes sparse LiDAR queries for effective multi-modal data fusion; (2) a Temporal Interaction and Update module that integrates temporally-predicted positions with current frame data, providing better initialization values for pose estimation and enhancing model's robustness against accumulative errors; and (3) a Temporal Clip Training strategy combined with a Collective Average Loss mechanism that aggregates losses across multiple frames, enabling global optimization and reducing the scale drift over long sequences.
Extensive experiments on the KITTI and Argoverse Odometry dataset demonstrate the superiority of our proposed DVLO4D, which achieves state-of-the-art performance in terms of both pose accuracy and robustness. Additionally, our method has high efficiency, with an inference time of 82 ms, possessing the potential for the real-time deployment.

\end{abstract}

\section{INTRODUCTION}

Visual/LiDAR odometry facilitates the estimation of relative pose transformations between consecutive image or point cloud frames.
It is a fundamental task in autonomous systems, including autonomous driving~\cite{wang2021pwclo, hu2023planning, shen2025oggaussianoccupancybasedstreet,cao2025real}, SLAM~\cite{deng2024compact,teed2024deep, xu2022multi, zhu2024sni}, and robotic localization ~\cite{liu2024toward, liu2024boosting,guan2024talk2radar,zhang2024detect}. 

Traditionally, odometry pipelines involve sequential stages of feature extraction, frame-to-frame matching, motion estimation, and optimization~\cite{graeter2018limo, huang2020lidar, wang2021dv, yuan2023sdv}. While effective, they frequently suffer from low accuracy due to suboptimal feature quality and resolution~\cite{wang2021dv}.
Learning-based approaches based on a single sensor have shown improved accuracy. However, they often face temporal misalignment or sensor degradation~\cite{deng2023long}. 
Visual-LiDAR odometry~\cite{wang2021dv, graeter2018limo, zhuoins20234drvo,liu2024dvlo}, integrating cameras and LiDAR sensors, has shown great potential by leveraging the strengths of both modalities. 
However, most of them~\cite{huang2020lidar,leutenegger2013keyframe,shin2020dvl,wang2021dv} cannot meet the real-time application, partly due to the complexity of processing data from multiple sensors. 
Additionally, the varying sensor configurations in different systems are often overlooked, which further complicates the development of efficient algorithms for various applications. 
Thus, a unified and efficient sensor fusion framework that can handle the specific demands of each environment is essential for advancing visual-LiDAR odometry research.
Moreover, incorporating temporal information alongside sensor fusion is critical for more robust feature tracking and matching.
However, this temporal information has been largely overlooked by existing multi-modal odometry methods which typically focus on pairwise frame inputs~\cite{liu2024dvlo, wang2022efficient,liu2023translo,wang2021pwclo}.

To overcome the above limitations, we propose DVLO4D, a query-based spatial-temporal fusion framework. 
Inspired by Deformable DETR~\cite{zhu2020deformable}, our approach leverages LiDAR features as queries, using their 3D positions as positional embeddings, and visual features as keys and values. 
By generating offsets around LiDAR reference points, the model adapts to focus on relevant regions in the image, capturing fine-grained details. 
This sparse query-based fusion method not only speeds up the computation compared to dense approaches, but also provides a unified framework that efficiently handles multi-modal data fusion. 
Additionally, we design a novel temporal fusion module, which is composed of a Memory Feature Bank (MFB) and a Memory Pose Bank (MPB). 
These banks, combined with a temporal encoder and cross-attention mechanism, allow the model to refine current pose estimates by leveraging historical cost volume features and previous pose estimates stored in MFB and MPB, respectively, leading to more accurate motion predictions.
Furthermore, to mitigate scale drift and improve overall accuracy,
we address the limitations of frame-to-frame loss optimization by introducing Temporal Clip Training and Collective Average Loss to optimize trajectory information globally. 
Overall, our contributions are as follows:
\begin{itemize}
\item We propose DVLO4D, a visual-LiDAR odometry with a sparse spatial-temporal fusion strategy. Our method utilizes both multi-modal and multi-frame information to improve the pose estimation accuracy and robustness.
\item A temporal fusion module is proposed to sufficiently leverage historical data assistance, combining an MFB and an MPB.
\item We also introduce Temporal Clip Training and Collective Average Loss to address the limitations of frame-to-frame loss.
\item We validate our method on KITTI and Argoverse dataset, achieving state-of-the-art performance in terms of both pose estimation accuracy and robustness.
\end{itemize}

\begin{figure*}[http]
\begin{center}
 \includegraphics[clip=true, trim=0in 0in 0in 0in, width=1\linewidth]{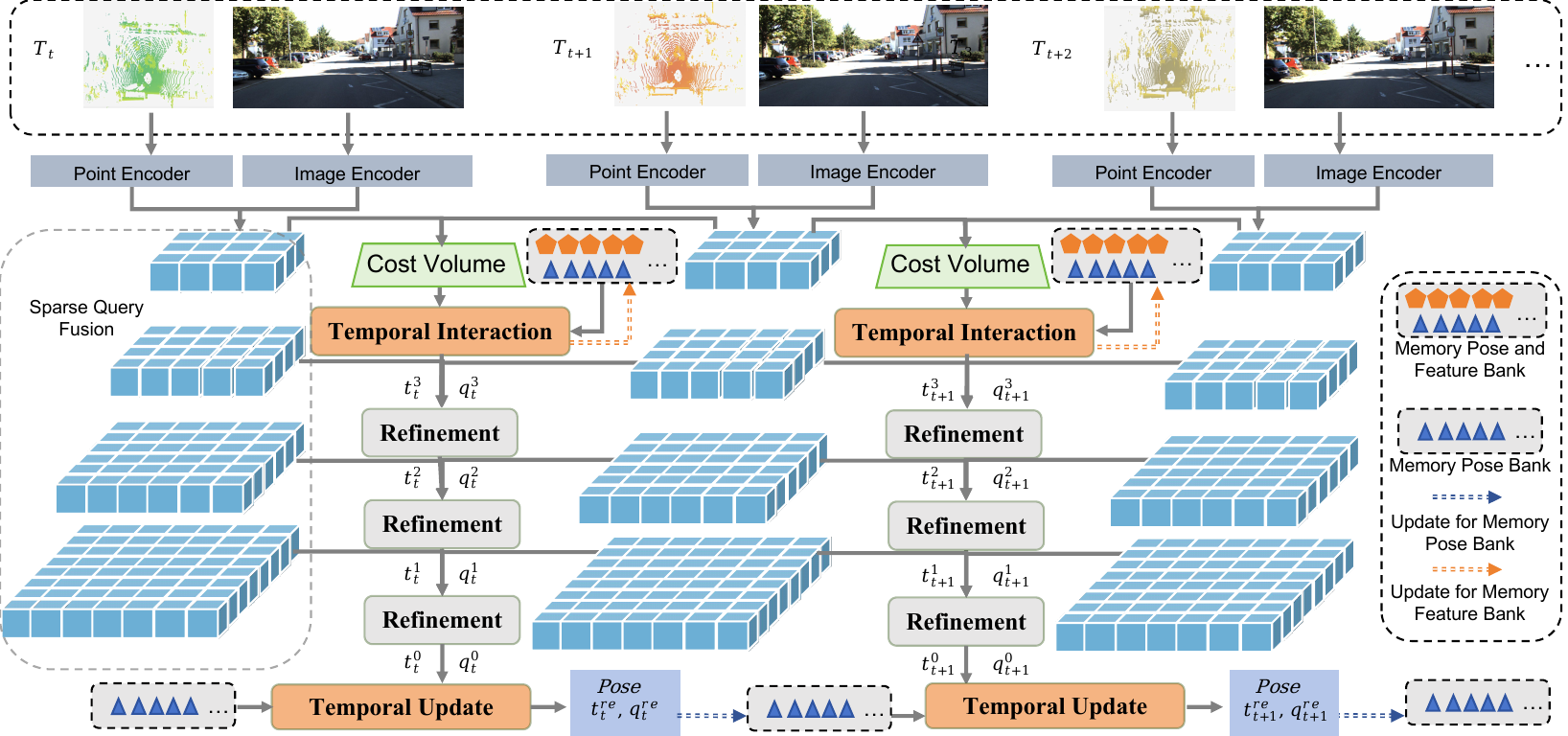} 
 \vspace{-0.5cm} 

\end{center}
  \caption{The pipeline of DVLO4D. The \textbf{Sparse Query Fusion (SQF)} mechanism utilizes sparse LiDAR queries to effectively fuse multi-modal data, enabling the integration of camera and LiDAR features. 
The \textbf{Temporal Interaction} and \textbf{Update (TIU)} module refine pose estimates by leveraging the Memory Feature Bank (MFB) and the Memory Pose Bank (MPB) and integrating them with the current input, improving the accuracy and robustness of the odometry estimation.}
\label{fig:framework}
\vspace{-0.5cm} 
\end{figure*}

\section{Related Work}
\subsection{Visual-LiDAR Odometry}
Visual-LiDAR odometry leverages the complementary strengths of visual and LiDAR systems, combining 2D texture and 3D geometric features to improve pose estimation. Existing approaches can be categorized into loosely integrated and tightly integrated methods. Loosely integrated methods~\cite{zhang2017real,graeter2018limo,shin2020dvl,huang2020lidar} use LiDAR data to enhance depth estimation and exploit visual tracking for pose estimation. For instance, DEMO~\cite{zhang2017real} assigns depth values \cite{xu2024sdge} to 2D feature points using LiDAR data, while LIMO~\cite{graeter2018limo} incorporates loop closure and semantic information. However, the lack of direct correspondences between 3D LiDAR points and 2D visual features often necessitates interpolation, which can introduce errors. 
Tightly integrated methods, such as V-LOAM~\cite{zhang2015visual}, aim for a more seamless fusion of visual and LiDAR data. On the learning-based front, methods like MVL-SLAM~\cite{an2022visual} fuse RGB images and LiDAR data using RCNNs, while LIP-Loc~\cite{shubodh2024lip} employs contrastive learning for cross-modal localization, though it struggles with structural differences between point clouds and images. Recent works such as~\cite{lai2022adafusion,liu2024dvlo} introduce descriptor fusion, and cluster techniques to improve feature integration.

Despite these advances, existing methods still struggle to generalize across different sensor configurations without extensive redesign. This underscores the need for a more flexible sensor fusion framework. To address this, we propose DVLO4D, which uses a query-based multi-modal fusion approach, providing a unified representation for diverse sensory inputs and improving the efficiency of feature fusion.

\subsection{Long-Term Odometry}

Recently, long-term visual odometry has gained significant attention, focusing on developing robust systems that maintain accuracy over extended periods and varying environments. DeepVO~\cite{wang2017deepvo} employs deep recurrent neural networks to model temporal dependencies in sequential data. Deng et al.~\cite{deng2023long} proposed a long-term SLAM system that incorporates map prediction and dynamics removal. 

Tracking Any Point (TAP) methods, such as PIPs~\cite{harley2022particle}, reformulate pixel tracking as a long-term motion estimation task \cite{liu2024difflow3d,jiang20243dsflabelling}, associating each pixel with a multi-frame trajectory. TAP-Net~\cite{doersch2022tap} builds on this with a global cost volume to refine correspondences, while BootsTAP~\cite{doersch2024bootstap} uses semi-supervised learning for further refinement. OmniMotion~\cite{wang2023tracking} introduces test-time optimization for point trajectories. LEAP-VO~\cite{chen2024leap} improves point tracking in dynamic scenes but focuses primarily on point-level correspondences, overlooking global temporal optimization.

In contrast, we propose a novel temporal odometry method that incorporates temporal interaction and update. Unlike most existing methods, which focus on frame-to-frame loss, our approach emphasizes global optimization across several frames. Our Collective Average Loss strategies globally optimize trajectory information, improving long-term odometry accuracy. Furthermore, our temporal interaction module simplifies network learning by providing a reliable initial pose estimate for each frame, which is then refined using temporal information. To the best of our knowledge, this is the first application of temporal interaction combined with collective average loss for odometry tasks, setting a new benchmark for long-term odometry performance.

\section{METHOD}

DVLO4D focuses on both multi-sensor (i.e., multi-view cameras and LiDAR) and multi-frame fusion for pose estimation, as shown in Fig.~\ref{fig:framework}.  Given different sensory inputs, we first apply modality-specific encoders to extract their multi-level image and point features. Then, we fuse multi-modal features using a unified query representation that preserves geometric and semantic information. Finally, the Temporal Prediction module, in collaboration with the MFB and MPB, provides a reliable initial pose estimate, which is refined iteratively. After the final iteration, an additional refinement step that incorporates temporal pose information is applied to further enhance the predicted pose accuracy. In the KITTI odometry task, the sensor inputs are two point clouds $PC_S, PC_T \in \mathbb{R}^{N \times 3}$ and their corresponding monocular camera images $I_S, I_T \in \mathbb{R}^{H \times W \times 3}$ from a pair of consecutive frames -- from the source frame $S$ to the target frame $T$, and the temporal inputs are MFB and MPB.

\subsection{Modality-Specific Encoder}

DVLO4D independently extracts features from LiDAR point clouds and multi-view camera images.

\noindent
\textbf{Point Encoder.} Due to the irregular and sparse nature of raw point clouds, following ~\cite{liu2023translo,liu2024dvlo,wang2022efficient}, we first project them onto a cylindrical surface to better organize the points. The 2D coordinates are computed by transforming the 3D coordinates $(x, y, z)$, where the horizontal position $u = \arctan2(y, x) / \Delta \theta$ and the vertical position $v = \arcsin(z / \sqrt{x^2 + y^2 + z^2}) / \Delta \phi$, with $\Delta \theta$ and $\Delta \phi$ representing the LiDAR sensor's horizontal and vertical resolutions, respectively~\cite{li2019net,liu2023regformer}. Each 2D position retains its original 3D coordinates, preserving the geometric information while converting the point cloud into a pseudo-image for multi-modal feature alignment. These pseudo-images, with dimensions $H_P \times W_P \times 3$, are fed into a hierarchical feature extraction module~\cite{wang2022efficient}, producing multi-level features $\mathbf{F}_P \in \mathbb{R}^{H_P \times W_P \times D}$, where $D$ is the number of channels. This approach effectively captures and utilizes the geometric information across multiple feature scales.

\noindent
\textbf{Image Encoder.} Given the input camera images $I \in \mathbb{R}^{H \times W \times 3}$, we employ a convolutional-based feature pyramid network~\cite{huang2020epnet} to extract image features $\mathbf{F}_I \in \mathbb{R}^{H \times W \times C}$, where $H$, $W$, and $C$ represent the height, width, and number of channels of the feature map, respectively. 


\subsection{Sparse Query Fusion}

We introduce a sparse query fusion mechanism inspired by Deformable DETR~\cite{zhu2020deformable}, where LiDAR features $\mathbf{F}_P$ serve as the initial query embeddings. These query embeddings are derived from multi-level point features and combined with the corresponding 3D coordinates of the LiDAR points, which act as positional embeddings. This allows the model to integrate and leverage features from multiple sensor modalities, including LiDAR, camera, and radar \cite{zhao2024unibevfusion,guan2023achelous++}.

Given a set of sparse LiDAR reference points $\mathbf{P} = \{P_i\}_{i=1}^{N}$~\cite{wang2022efficient}, where each $P_i \in \mathbb{R}^3$ represents the 3D position of a LiDAR point, we project them onto the 2D image plane using camera intrinsic and extrinsic parameters. These positions in the $k$-th camera view are denoted as $\mathbf{T}_k(P_i)$. To enhance the feature extraction, adaptive offsets $\Delta \mathbf{u}_\text{cam}^{ikj}$ and attention weights $\alpha_\text{cam}^{ikj}$ are generated around each projected point, creating multiple sampling locations in the image space. The sampled visual features $\mathbf{F}_\text{sample}^{i}$ are calculated as:
\vspace{-6pt}
\begin{equation}
\mathbf{F}_\text{sample}^{i} = \sum_{k=1}^{N_c} \sum_{j=1}^{M} \mathbf{F}_\text{cam}^{k}(\mathbf{T}_k(P_i) + \Delta \mathbf{u}_\text{cam}^{ikj}) \cdot \alpha_\text{cam}^{ikj},
\vspace{-3pt}
\end{equation}
where $\mathbf{F}_\text{cam}^{k}$ is the bilinear sampling of features from the $k$-th camera's feature map at the offset location $(\mathbf{T}_k(P_i) + \Delta \mathbf{u}_\text{cam}^{ikj})$, $N_c$ denotes the number of camera views and $M$ is the number of sampling locations around the $i$-th point.

These sampled image features are then used in a multi-head cross-attention (MHCA) mechanism, where the LiDAR features (queries) are fused with the sampled visual features (keys and values). The cross-attention is computed as:
\vspace{-6pt}
\begin{equation}
\label{eq:fusion}
\mathbf{F}_{\text{attn}} = \text{MHCA}(\mathbf{F}_P, \mathbf{F}_\text{sample}, \mathbf{F}_\text{sample}),
\vspace{-3pt}
\end{equation}

where $\mathbf{F}_{\text{cam}}$ acts as both the key and value. This effectively fuses the sparse LiDAR data with the rich visual context, enhancing the overall feature representation. Finally, to address the limited receptive field in sampling fusion, we adopt a global adaptive fusion mechanism from~\cite{peng2023delflow,liu2024dvlo}. 

The hierarchical features (Fig.~\ref{fig:framework} in blue cubes) represent the multi-level fused features by fusing LiDAR and image features. Eq.~\eqref{eq:fusion} is applied across all feature levels to ensure consistent feature fusion. To project LiDAR points onto varying feature map levels, we normalize the LiDAR coordinates, ensuring consistent alignment with image features.




\subsection{Iterative Pose Estimation}


Following the method in \cite{wang2022efficient,wang2021pwclo}, we use an attentive cost volume to derive coarse embedding features \( E \in \mathbb{R}^{N \times D} \) by correlating fused features \( F_{S}^{L-1} \) and \( F_{T}^{L-1} \) from consecutive frames in the coarsest layer, where \( L \) is the total number of layers. These features encapsulate inter-frame correlations.

The quaternion \( q \in \mathbb{R}^4 \) and translation vector \( t \in \mathbb{R}^3 \) are computed as \( q = \frac{\text{FC}(E \odot M)}{\|\text{FC}(E \odot M)\|} \) and \( t = \text{FC}(E \odot M) \), where \( M \in \mathbb{R}^{N \times D} \) represents learnable masks. The initial estimates are refined iteratively using \cite{wang2021pwclo}. The refined quaternion \( q^l \) and translation \( t^l \) at layer \( l \) are updated as \( q^l = \Delta q^l q^{l+1} \) and \( \left[0, t^l\right] = \Delta q^l\left[0, t^{l+1}\right]\left(\Delta q^l\right)^{-1} + \left[0, \Delta t^l\right] \), where residuals \( \Delta q^l \) and \( \Delta t^l \) follow a similar process as in the coarsest layer.


\begin{figure}[t] 
\begin{center}
 \includegraphics[clip=true, trim=0in 1.8cm 0in 0in, width=1\linewidth]{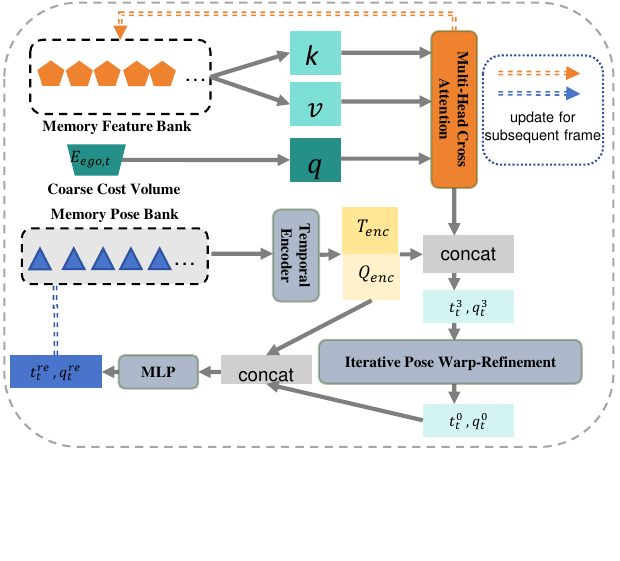} 
 \vspace{-0.5cm} 
\end{center}
  \caption{Temporal Interaction and Update module.}
\label{fig:TIU}
\vspace{-0.5cm} 
\end{figure}


\subsection{Temporal Interaction and Update Module}

\noindent
\textbf{Temporal Interaction Module.}
In the Temporal Interaction and Update Module, the current pose estimation is refined by temporal information from past frames, as detailed in Fig.~\ref{fig:TIU}. This module uses the MFB and MPB to store historical fused features and pose estimates.

At initialization (T=0), both memory banks are set to zero vectors. As new frames are processed (T=1, T=2, ..., T=N), the most recent feature and pose estimates are added to the MFB and MPB. When the number of frames exceeds a threshold, older frames are discarded. This sliding window mechanism ensures efficient memory usage, focusing on the most recent context.

To refine the pose, we initialize the ego instance feature \( \mathbf{E}_{\text{ego}} \) using the coarsest cost volume feature map $L-1$. It provides two key benefits: the coarsest feature map encapsulates the semantic and geometric context of the driving scene, and its dense representation complements sparse representation, particularly in handling dynamic objects and occlusions.

For the historical quaternions and translations, we maintain a sequence of refined poses across \( T_h \) frames in the MPB, denoted as \( \mathbf{q}_{t_h} \in \mathbb{R}^{T_h \times 4} \) and \( \mathbf{t}_{t_h} \in \mathbb{R}^{T_h \times 3} \), where $T_h$ stands for the past time horizon. These are processed using a temporal encoder consisting of a Transformer followed by an LSTM,  separately. Concretely, the embeddings \( \mathbf{Q}_{\text{enc}} \) and \( \mathbf{T}_{\text{enc}} \) are produced by applying Multi-Head Self-Attention (MHSA) to the historical quaternions and translations, followed by the LSTM to capture the temporal dynamics.

The ego feature \( \mathbf{E}_{\text{ego}} \) is refined using another MHCA by Eq. \eqref{eq:MHCA}, which attends to the features stored in the MFB.
\vspace{-6pt}
\begin{equation}
\label{eq:MHCA}
\mathbf{E}_{\text{ego}} = \text{MHCA}(\mathbf{E}_{\text{ego}, t_1}, \mathbf{E}_{\text{ego}, t_h}, \mathbf{E}_{\text{ego}, t_h}),
\vspace{-6pt}
\end{equation}
where the features from all \( T_h \) frames are used to refine the \( \mathbf{E}_{\text{ego}, t_1} \) at the current frame $t_1$. The embeddings \( \mathbf{Q}_{\text{enc}} \) and \( \mathbf{T}_{\text{enc}} \) are then concatenated with \( \mathbf{E}_{\text{ego}} \), and passed through an MLP to predict the first layer quaternion 
\vspace{-6pt}
\begin{align} 
{q}^{L-1} &= \frac{\text{MLP}(\text{Concat}(\mathbf{E}_{\text{ego}},  \mathbf{Q}_{\text{enc}}))}{\|\text{MLP}(\text{Concat}(\mathbf{E}_{\text{ego}},  \mathbf{Q}_{\text{enc}}))\|}, \\
 {t}^{L-1} &= \text{MLP}(\text{Concat}(\mathbf{E}_{\text{ego}}, \mathbf{T}_{\text{enc}})). 
 \vspace{-3pt}
\end{align}

\noindent
\textbf{Temporal Update Module.}  
The Update Module re-encodes the last layer iterrative predictions \( {q}^{0} \) and \( {t}^{0} \) using an MLP. These encoded features \( \mathbf{F}_{{q}^{0}} \) and \( \mathbf{F}_{{t}^{0}} \) are concatenated with the historical features \( \mathbf{Q}_{\text{enc}} \) and \( \mathbf{T}_{\text{enc}} \) from the MPB to produce the final refined pose: 
\vspace{-6pt}
\begin{align} 
{q}^\text{re} &= \frac{\text{MLP}(\text{Concat}(\mathbf{F}_{{q}^{0}},  \mathbf{Q}_{\text{enc}}))}{\|\text{MLP}(\text{Concat}(\mathbf{F}_{{q}^{0}},  \mathbf{Q}_{\text{enc}}))\|},\\
{t}^\text{re} &= \text{MLP}(\text{Concat}(\mathbf{F}_{\mathbf{t}^{0}}, \mathbf{T}_{\text{enc}})).
\end{align}

By incorporating historical pose and feature information from the MPB and MFB, our model provides a more accurate initialization and refined pose estimation for each frame.

\subsection{Collective Averge Loss}

Our network outputs pose estimates \( q^l \) and \( t^l \) from four layers, which are used to compute the supervised loss \( \mathcal{L}^l \) as described in \cite{wang2022efficient, liu2024dvlo}. Concisely, for the \( l \)-th layer:
\vspace{-3pt}
{\small
\begin{equation}
\mathcal{L}^l = \| t_{\text{gt}} - t^l \| \exp(-k_t) + k_t + \| q_{\text{gt}} - q^l \|_2 \exp(-k_q) + k_q,
\end{equation}
}
where \( t_{\text{gt}} \) and \( q_{\text{gt}} \) are the ground truth translation and quaternion, respectively. The terms \( k_t \) and \( k_q \) are learnable scalars that scale the losses. The \( L_1 \) and \( L_2 \) norms are used for translation and rotation losses.
The loss for the refined predictions \( {t}^\text{re} \) and \( {q}^\text{re} \) is similarly defined as:
\vspace{-3pt}
{\small
\begin{equation}
\mathcal{L}_{\text{re}} = \| t_{\text{gt}} - {t}^\text{re} \| \exp(-k_t) + k_t + \| q_{\text{gt}} - {q}^\text{re} \|_2 \exp(-k_q) + k_q.
\end{equation}
}

Training samples play a crucial role in the temporal modeling of odometry, since DVLO4D learns temporal variations from the data itself, rather than relying on manually designed statistically optimal estimators such as Kalman Filtering~\cite{kalman1960new}. Hence, conventional pairwise training methods are insufficient for generating training samples necessary for long-range odometry estimation. In contrast, DVLO4D processes video clips as input, enabling the generation of training samples that support temporal learning over extended sequences.

To further improve temporal learning, inspired by MOTR~\cite{zeng2022motr}, we introduce the collective average loss (CAL), which aggregates losses over multiple frames. CAL is computed as the total loss over the sub-clip frames \( T_s \):
\vspace{-6pt}
\begin{equation}
\mathcal{L}_{\text{CAL}} = \frac{1}{T_s} \sum_{t=1}^{T_s} \left( \sum_{l=1}^{L} \alpha^l \mathcal{L}^l_t + \beta \mathcal{L}_{\text{re},t} \right),
\vspace{-3pt}
\end{equation}
where \( \mathcal{L}^l_t \) is the loss for the \( l \)-th layer at frame \( t \), \( \alpha^l \) and \( \beta \) are weights for layer losses and refined loss.


\begin{table*}[t]
\centering
\caption{Comparison with different odometry methods on the KITTI odometry dataset. $t_{\text{rel}}$ and $r_{\text{rel}}$ represent the average sequence translational RMSE (\%) and sequence rotational RMSE (°/100m) respectively in the length of 100, 200, ..., 800m. The best results are \textcolor{red}{bold}, and the second best results are \textcolor{blue}{underlined}. * represents the model trained on the 00-08 sequences.}
\label{table:kitti_comparison}
{\renewcommand{\arraystretch}{1.1}
\setlength{\tabcolsep}{3.3pt}
\resizebox{\textwidth}{!}{
\fontsize{8pt}{8pt}\selectfont
    \begin{tabular}{l|c c|c c|c c|c c|c c|c c|c c|c c|c c|c c|c c|c c c}
    \hline
    \multirow{2}{*}{\textbf{Method}} & \multicolumn{2}{c|}{00} & \multicolumn{2}{c|}{01} & \multicolumn{2}{c|}{02} & \multicolumn{2}{c|}{03} & \multicolumn{2}{c|}{04} & \multicolumn{2}{c|}{05} & \multicolumn{2}{c|}{06} & \multicolumn{2}{c|}{07} & \multicolumn{2}{c|}{08} & \multicolumn{2}{c|}{09} & \multicolumn{2}{c|}{10} & \multicolumn{2}{c}{mean (07-10)} \\
    \cline{2-25}
    & \(t_{\text{rel}}\) & \(r_{\text{rel}}\) & \(t_{\text{rel}}\) & \(r_{\text{rel}}\) & \(t_{\text{rel}}\) & \(r_{\text{rel}}\) & \(t_{\text{rel}}\) & \(r_{\text{rel}}\) & \(t_{\text{rel}}\) & \(r_{\text{rel}}\) & \(t_{\text{rel}}\) & \(r_{\text{rel}}\) & \(t_{\text{rel}}\) & \(r_{\text{rel}}\) & \(t_{\text{rel}}\) & \(r_{\text{rel}}\) & \(t_{\text{rel}}\) & \(r_{\text{rel}}\) & \(t_{\text{rel}}\) & \(r_{\text{rel}}\) & \(t_{\text{rel}}\) & \(r_{\text{rel}}\) & \(t_{\text{rel}}\) & \(r_{\text{rel}}\) \\
    \hline
    \multicolumn{25}{l}{\textbf{Visual Odometry Methods}} \\
    \hline
    SfMLearner\textsuperscript{*}~\cite{zhou2017unsupervised} & 21.32 & 6.19 & 22.41 & 2.79 & 24.10 & 4.18 & 12.56 & 4.52 & 4.32 & 3.28 & 12.59 & 4.66 & 15.55 & 5.58 & 12.61 & 6.31 & 10.66 & 3.75 & 11.32 & 4.07 & 15.25 & 4.06 & 12.46 & 4.55 \\
TartanVO~\cite{wang2021tartanvo}& -- & -- &-- & --  &-- & -- &--& -- &-- & -- &-- & -- & 4.72& 2.95& 4.32& 3.41&-- & -- &6.0 &3.11& 6.89 &2.73& 5.48& 3.05 \\
    MLM-SFM\textsuperscript{*}~\cite{song2015high} & 2.04 & 0.48 & -- & -- & 1.50 & 0.35 & 3.37 & \textcolor{red}{\textbf{0.21}} & 1.43 & 0.23 & 2.19 & 0.38 & 2.09 & 0.81 & -- & -- & 2.37 & 0.44 & 1.76 & 0.47 & 2.12 & 0.85 & 2.08 & 0.59 \\
    DFVO\textsuperscript{*}~\cite{zhan2021df} & 2.01 & 0.79 & 61.17 & 18.96 & 2.46 & 0.79 & 3.27 & 0.89 & 0.79 & 0.56 & 1.50 & 0.74 & 1.95 & 0.76 & 2.28 & 1.16 & 2.11 & 0.74 & 3.21 & 0.59 & 2.89 & 0.97 & 2.62 & 0.87 \\
    Cho et al.\textsuperscript{*}~\cite{cho2023dynamic} & 1.77 & 0.79 & 64.38 & 16.87 & 2.62 & 0.74 & 3.06 & 0.89 & 0.65 & 0.55 & 1.31 & 0.74 & 1.60 & 0.56 & 1.06 & 0.67 & 2.28 & 0.76 & 2.66 & 0.53 & 2.95 & 0.95 & 2.24 & 0.73 \\
    \hline
    \multicolumn{25}{l}{\textbf{LiDAR Odometry Methods}} \\
    \hline
    LO-Net~\cite{li2019net} & 1.47 & 0.72 & 1.36 & 0.47 & 1.52 & 0.71 & 1.03 & 0.66 & 0.51 & 0.65 & 1.04 & 0.69 & 0.71 & 0.50 & 1.70 & 0.89 & 2.12 & 0.77 & 1.37 & 0.58 & 1.80 & 0.93 & 1.75 & 0.79 \\
    PWCLO~\cite{wang2021pwclo} & 0.89 & 0.43 & 1.11 & 0.42 & 1.87 & 0.76 & 1.42 & 0.92 & 1.15 &0.94 &1.34 &0.71 &0.60 &0.38 &1.16 &1.00 &1.68 &0.72 &0.88& 0.46 &2.14 &0.71& 1.47 &0.72 \\
    DELO~\cite{ali2023delo} & 1.43 & 0.81 & 2.19 & 0.57 & 1.48 & 0.52 &  1.38 &1.10 &2.45& 1.70 &1.27 &0.64 &0.83 &0.35 &0.58 &0.41 &1.36 &0.64 &1.23 &0.57 &1.53 &0.90 &1.18 &0.63\\
    TransLO~\cite{liu2023translo} & 0.85 & 0.38 & 1.16 & 0.45 & 0.88 & 0.34 &  1.00 &0.71 &0.34 &\textcolor{blue}{\underline{0.18}}& 0.63 &0.41 &0.73 &0.31 &0.55& 0.43 &1.29 &0.50 &0.95 &0.46 &1.18 &0.61 &0.99 &0.50 \\
    EfficientLO~\cite{wang2022efficient} & \textcolor{blue}{\underline{0.80}} & 0.37 & 0.91 & 0.40 & 0.94 & 0.32 & \textcolor{blue}{\underline{0.51}} & 0.43 & 0.38 & 0.30 & 0.57 & 0.33 & 0.36 & 0.23 & \textcolor{red}{\textbf{0.37}} & \textcolor{red}{\textbf{0.26}} & 1.22 & 0.48 & 0.87 & 0.38 & 0.91 & \textcolor{blue}{\underline{0.50}} & 0.86 & \textcolor{blue}{\underline{0.41}} \\
    \hline
    \multicolumn{25}{l}{\textbf{Multimodal Odometry Methods}} \\\hline
An et al.\textsuperscript{*}~\cite{an2022visual} & 2.53 & 0.79 & 3.76 & 0.80 & 3.95 & 1.05 & 2.75 & 1.39 & 1.81 & 1.48 & 3.49 & 0.79 &1.84& 0.83& 3.27 & 1.51 & 2.75 & 1.61 & 3.70 & 1.83 & 4.65 & 0.51 & 3.59 & 1.37 \\
H-VLO\textsuperscript{*}~\cite{aydemir2022h} & 1.75 & 0.62 & 4.32 & 0.46 & 2.32 & 0.60 &  2.52 &0.47 &0.73 &0.36 &0.85 &0.35&0.75 &0.30 &0.79 &0.48 &1.35& \textcolor{blue}{\underline{0.38}} &1.89& \textcolor{blue}{\underline{0.34}} & 1.39 &0.52 &1.36 &0.43 \\
DVLO~\cite{liu2024dvlo} & \textcolor{blue}{\underline{0.80}} & \textcolor{blue}{\underline{0.35}} & \textcolor{blue}{\underline{0.85}} & \textcolor{blue}{\underline{0.33}} & \textcolor{blue}{\underline{0.81}} & \textcolor{red}{\textbf{0.29}} & 0.59 & 0.36 & \textcolor{blue}{\underline{0.26}} & \textcolor{red}{\textbf{0.13}} & \textcolor{blue}{\underline{0.41}} & \textcolor{blue}{\underline{0.23}} & \textcolor{blue}{\underline{0.33}} & \textcolor{red}{\textbf{0.17}} &  0.46 &0.33& \textcolor{blue}{\underline{1.09}} & 0.44 &\textcolor{blue}{\underline{0.85}} &0.36 &\textcolor{blue}{\underline{0.88}} &\textcolor{red}{\textbf{0.46}} & \textcolor{blue}{\underline{0.82}} & \textcolor{blue}{\underline{0.41}} \\
\textbf{DVLO4D (Ours)} & \textcolor{red}{\textbf{0.68}} & \textcolor{red}{\textbf{0.33}} &\textcolor{red}{\textbf{0.77}} & \textcolor{red}{\textbf{0.23}}  & \textcolor{red}{\textbf{0.76}} & \textcolor{blue}{\underline{0.31}} & \textcolor{red}{\textbf{0.49}} & \textcolor{blue}{\underline{0.33}} &\textcolor{red}{\textbf{0.22}} & \textcolor{red}{\textbf{0.13}} &\textcolor{red}{\textbf{0.39}} & \textcolor{red}{\textbf{0.21}} & \textcolor{red}{\textbf{0.32}} & \textcolor{blue}{\underline{0.21}} &\textcolor{blue}{\underline{0.43}} & \textcolor{blue}{\underline{0.32}} & \textcolor{red}{\textbf{0.95}} & \textcolor{red}{\textbf{0.36}} & \textcolor{red}{\textbf{0.77}} & \textcolor{red}{\textbf{0.33}} & \textcolor{red}{\textbf{0.76}} & \textcolor{red}{\textbf{0.46}} & \textcolor{red}{\textbf{0.73}} & \textcolor{red}{\textbf{0.37}}  \\

\hline
\end{tabular}}}
 \vspace{-3mm}
\end{table*}


\begin{table}[t]
    \centering
        \caption{\textbf{Experiments on the Argoverse dataset~\cite{argo}.}}
        {
        \footnotesize
        \vspace{-6pt}
        \tabcolsep 3pt
        \renewcommand{\arraystretch}{1.2}
        \fontsize{5pt}{5pt}\selectfont
        \resizebox{0.48\textwidth}{!}{ 
        \begin{tabular}{p{4cm}|c|c}
            \hline
            \multirow{2}{*}{\textbf{Method}} & \multicolumn{2}{c}{\textbf{Mean on 00-23}} \\
            \cline{2-3} & \textbf{ATE} & \textbf{RPE} \\
            \hline
            PyLiDAR~\cite{pylidar} w/o mapping & 6.900 & 0.109 \\
            A-LOAM~\cite{ALOAM} w/o mapping & 4.138 & 0.066 \\
            DSLO~\cite{zhang2024dslo} & \textcolor{blue}{\underline{0.111}} & \textcolor{blue}{\underline{0.027}} \\
            \hline
            {\textbf{DVLO4D (Ours)}} & \textcolor{red}{\textbf{0.089}} & \textcolor{red}{\textbf{0.025}} \\
            \hline
        \end{tabular}
        \label{tab:argo-acc}    
    }}
    \vspace{-6pt}
\end{table}

\begin{table}[h]
    \centering
    \caption{Comparison with learning-based multi-modal odometry on KITTI 09-10 sequences. Our DVLO4D is trained on 00-06 sequences while other models are trained on 00-08 sequences. The best results are \textcolor{red}{bold}.}
    \label{tab:learning_based_comparison}
    {\renewcommand{\arraystretch}{1.1}
    \fontsize{8pt}{8pt}\selectfont
    \resizebox{0.485\textwidth}{!}{ 
    \begin{tabular}{l|c|c|c|c|c|c|c}
        \hline
        \multicolumn{1}{c|}{\multirow{2}{*}{\textbf{Method}}} & \multicolumn{1}{c|}{\multirow{2}{*}{\textbf{Modalities}}} & \multicolumn{2}{c|}{\textbf{09}} & \multicolumn{2}{c|}{\textbf{10}} & \multicolumn{2}{c}{\textbf{Mean (09-10)}} \\
        \cline{3-8}
         &  & $t_{rel}$ & $r_{rel}$ & $t_{rel}$ & $r_{rel}$ & $t_{rel}$ & $r_{rel}$ \\
        \hline
        Self-VIO~\cite{li2021self} & visual+LiDAR & 2.58 & 1.13 & 2.67 & 1.28 & 2.62 & 1.21 \\
        SelfVIO~\cite{almalioglu2022selfvio} & visual+LiDAR & 1.95 & 1.15 & 1.81 & 1.30 & 1.88 & 1.23 \\
        VIOLearner~\cite{shamwell2019unsupervised} & visual+inertial & 1.82 & 1.08 & 1.74 & 1.38 & 1.78 & 1.23 \\
        H-VLO~\cite{aydemir2022h} & visual+LiDAR & 1.89 & 0.34 & 1.39 & 0.52 & 1.64 & 0.43 \\
        DVLO4D (Ours) & visual+LiDAR &\textcolor{red}{\textbf{0.77}} &\textcolor{red}{\textbf{0.33}}  &\textcolor{red}{\textbf{0.76}} & \textcolor{red}{\textbf{0.46}}  & \textcolor{red}{\textbf{0.77}} & \textcolor{red}{\textbf{0.40}} \\
        \hline
    \end{tabular}
    }}
    \vspace{-3mm}
\end{table}


\begin{table}[h]
  \centering
  \caption{Average inference time of different multi-modal odometry methods on the sequence 07-10 of KITTI dataset.}
  \resizebox{1\linewidth}{!}{
  \setlength{\tabcolsep}{4.5pt}
  \fontsize{8pt}{8pt}\selectfont
  {\renewcommand{\arraystretch}{1.1}
  \begin{tabular}{c|c|c|c|c|c|c}
    \hline
    \multirow{2}{*}{\textbf{Method}} & DV-LOAM & PL-LOAM & OKVIS-S & Shu \emph{et al.} & DVLO & DVLO4D \\
     & \cite{wang2021dv} & \cite{huang2020lidar} & \cite{leutenegger2013keyframe} & ~\cite{shu2022multi} & \cite{liu2024dvlo} & (Ours) \\
    \hline
    Time (ms)& 167  & 200 & 143 & 100 & 98.5 & 82 \\
    \hline
  \end{tabular}}}
  \label{table:inference_time}
\end{table}

\begin{table}[ht]
\centering
\caption{Robustness comparison under different settings. ``5Hz": downsampling LiDAR to half the original frequency. ``Noise": adding Gaussian noise to LiDAR data.}
\label{tab:robustness}
\scriptsize  
{\renewcommand{\arraystretch}{1.2}
\setlength{\tabcolsep}{1.5pt}
\resizebox{1\linewidth}{!}{%
\fontsize{8pt}{8pt}\selectfont
\begin{tabular}{c|c|cc|cc|cc|cc|cc}
\hline
\multirow{2}{*}{\textbf{Method}} & \multirow{2}{*}{\textbf{Setting}} & \multicolumn{2}{c|}{\textbf{07}} & \multicolumn{2}{c|}{\textbf{08}} & \multicolumn{2}{c|}{\textbf{09}} & \multicolumn{2}{c|}{\textbf{10}} & \multicolumn{2}{c}{\textbf{Mean}} \\ \cline{3-12} 
 &  & \textit{t\textsubscript{rel}} & \textit{r\textsubscript{rel}} & \textit{t\textsubscript{rel}} & \textit{r\textsubscript{rel}} & \textit{t\textsubscript{rel}} & \textit{r\textsubscript{rel}} & \textit{t\textsubscript{rel}} & \textit{r\textsubscript{rel}} & \textit{t\textsubscript{rel}} & \textit{r\textsubscript{rel}} \\ \hline
EfficientLO\cite{wang2022efficient} & \multirow{2}{*}{10HZ} & 0.37 & 0.26 & 1.22 & 0.48 & 0.87 & 0.38 & 0.91 & 0.50 & 0.86 & 0.41 \\ \cline{3-12} \cline{1-1}
DVLO4D(Ours)  & & 0.43 &0.32 &0.95 &0.36 &0.77 &0.33 &0.76 &0.46 &0.73 &0.37 \\ \hline
EfficientLO\cite{wang2022efficient} &  \multirow{2}{*}{5HZ} & 0.45 & 0.35 & 1.42 &0.57 & 1.03 & 0.44 & 1.04 & 0.60 & 0.99\textcolor{red}{($\uparrow$15\%)} & 0.49\textcolor{red}{($\uparrow$20\%)} \\ \cline{3-12} \cline{1-1}
DVLO4D(Ours)  & & 0.45 & 0.33 & 1.03 & 0.41 & 0.73 & 0.35 & 0.78 & 0.47 & 0.75\textcolor{red}{($\uparrow$3\%)} & 0.39\textcolor{red}{($\uparrow$5\%)} \\ \hline
EfficientLO\cite{wang2022efficient} & Add & 0.42 & 0.30 & 1.36 & 0.55 & 0.96 & 0.45 & 1.13 & 0.53 & 0.97\textcolor{red}{($\uparrow$13\%)} & 0.46\textcolor{red}{($\uparrow$12\%)} \\ \cline{3-12} \cline{1-1}

DVLO4D(Ours)  & noise & 0.44 & 0.32 & 0.98 & 0.37 & 0.79 & 0.35 & 0.79 & 0.46 & 0.75\textcolor{red}{($\uparrow$3\%)} & 0.37\textcolor{red}{($\uparrow$0\%)} \\ \hline
\end{tabular}%
}}
\vspace{-3mm}
\end{table}


\begin{table}[h]
  \centering
  \caption{Significance of SQF (Sparse Query Fusion), TIU (Temporal Interaction and Update module), and CAL (Collective Averge Loss) in DVLO4D network.}
  {\renewcommand{\arraystretch}{1.1}
  \setlength{\tabcolsep}{4pt}
  \resizebox{1\linewidth}{!}{
  \fontsize{8pt}{8pt}\selectfont
  \begin{tabular}{c|c|c|cc|cc|cc|cc|cccc}
    \hline
    \multirow{2}{*}{SQF} & \multirow{2}{*}{TIU} & \multirow{2}{*}{CAL} & \multicolumn{2}{c|}{07} & \multicolumn{2}{c|}{08} & \multicolumn{2}{c|}{09} & \multicolumn{2}{c|}{10} & \multicolumn{2}{c}{Mean (07-10)} \\ 
    \cline{4-13}
    & & & \emph{t\(_{rel}\)} & \emph{r\(_{rel}\)} & \emph{t\(_{rel}\)} & \emph{r\(_{rel}\)} & \emph{t\(_{rel}\)} & \emph{r\(_{rel}\)} & \emph{t\(_{rel}\)} & \emph{r\(_{rel}\)} & \emph{t\(_{rel}\)} & \emph{r\(_{rel}\)} \\
    \hline
    \checkmark & \checkmark &  & 0.47 & 0.39 & 1.03 & 0.41 & 0.98 & 0.52 & 0.83 & 0.64 & 0.83 & 0.49 \\
    \checkmark &  & \checkmark & 0.55 & 0.41 & 1.08 & 0.51 & 0.95 & 0.55 & 0.92 & 0.68 & 0.92 & 0.54 \\
    & \checkmark & \checkmark & 0.53 & 0.48 & 1.20 & 0.56 & 1.01 & 0.55 & 0.96 & 0.67 & 0.98 & 0.57 \\
    \checkmark & \checkmark & \checkmark & \textbf{0.43} & \textbf{0.32} & \textbf{0.95} & \textbf{0.36} & \textbf{0.77} & \textbf{0.33} & \textbf{0.76} & \textbf{0.46} & \textbf{0.73} & \textbf{0.37} \\
    \hline
  \end{tabular}}}
  \label{table:ablation_tiu_sqf_cal}
   \vspace{-3mm}
\end{table}

\begin{figure}[t]
    \centering
    
    \begin{subfigure}[b]{0.45\linewidth}
        \centering
        \includegraphics[width=\textwidth, trim=30 35 0 60, clip]{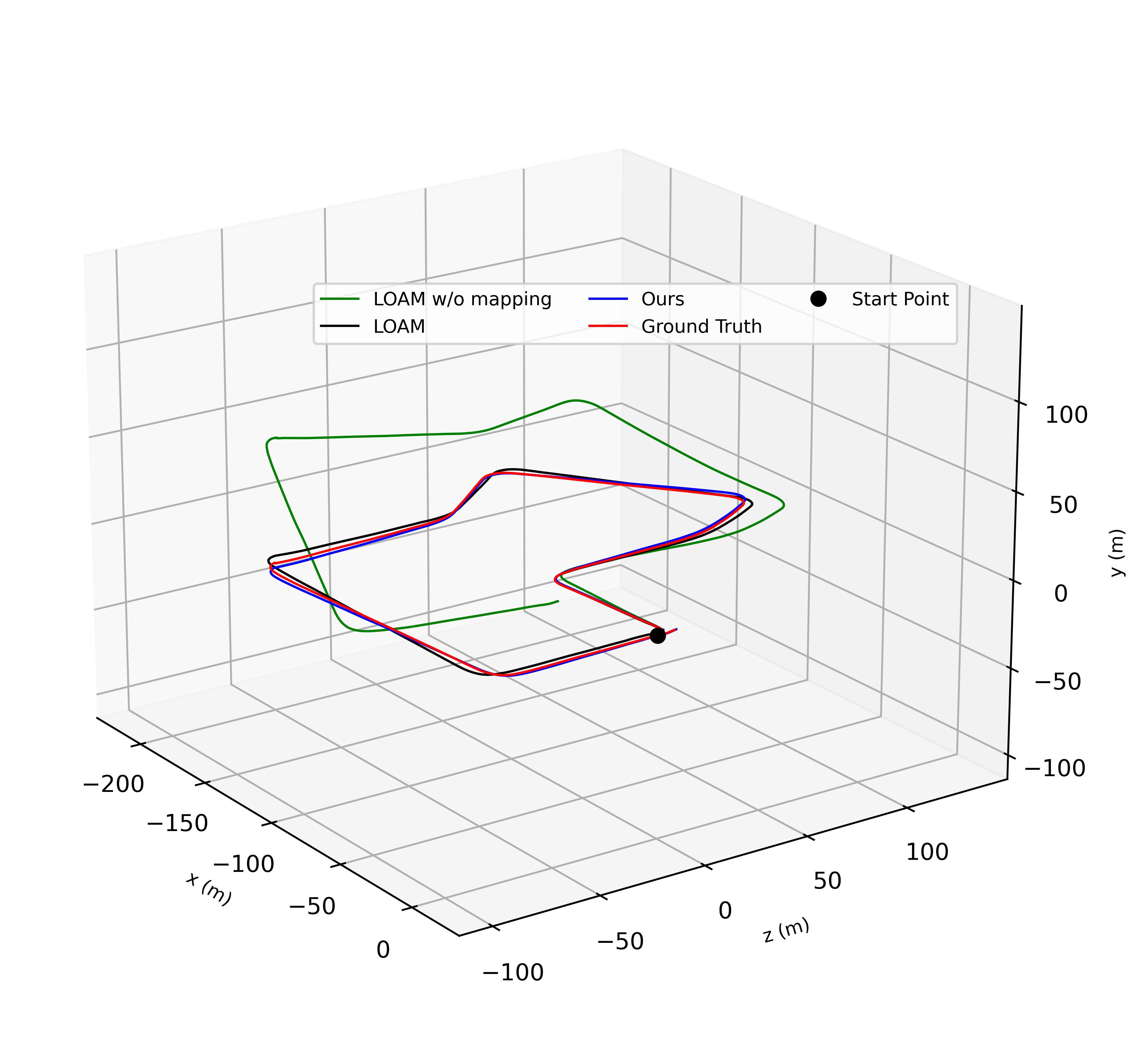}
        \caption{3D trajectory of seq.07}
    \end{subfigure}
    \begin{subfigure}[b]{0.45\linewidth}
        \centering
        \includegraphics[width=\textwidth, trim=30 35 0 60, clip]{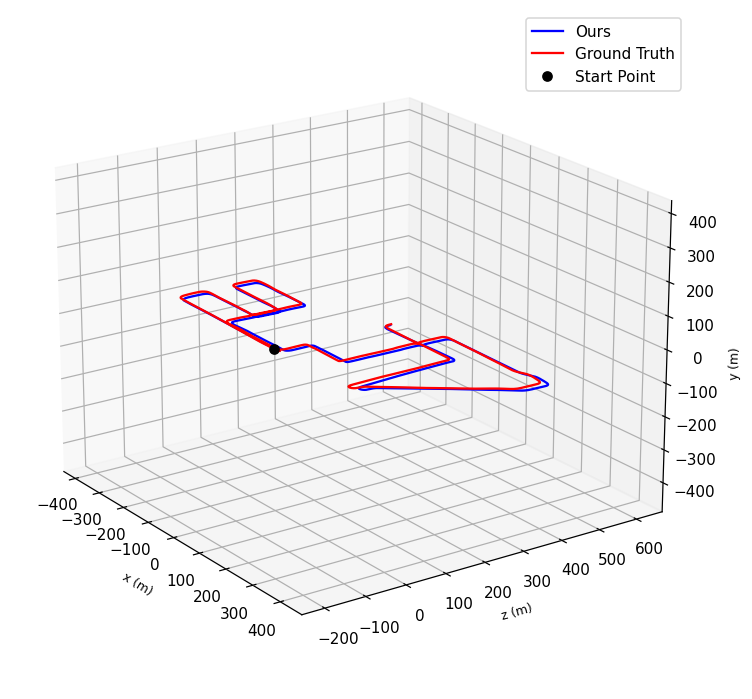}
        \caption{3D trajectory of seq.08}
    \end{subfigure}
    
    \begin{subfigure}[b]{0.45\linewidth}
        \centering
        \includegraphics[width=\textwidth, trim=30 35 0 60, clip]{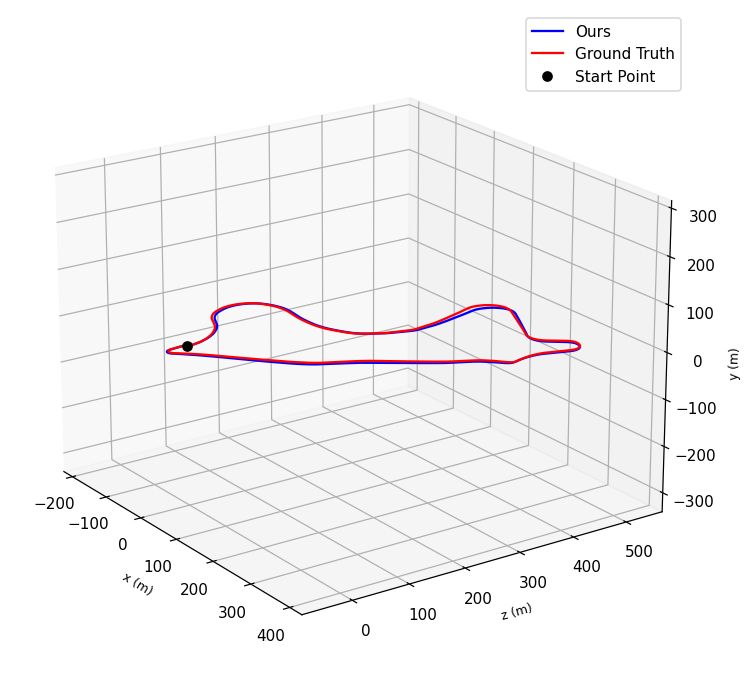}
        \caption{3D trajectory of seq.09}
    \end{subfigure}
    \begin{subfigure}[b]{0.45\linewidth}
        \centering
        \includegraphics[width=\textwidth, trim=30 35 0 60, clip]{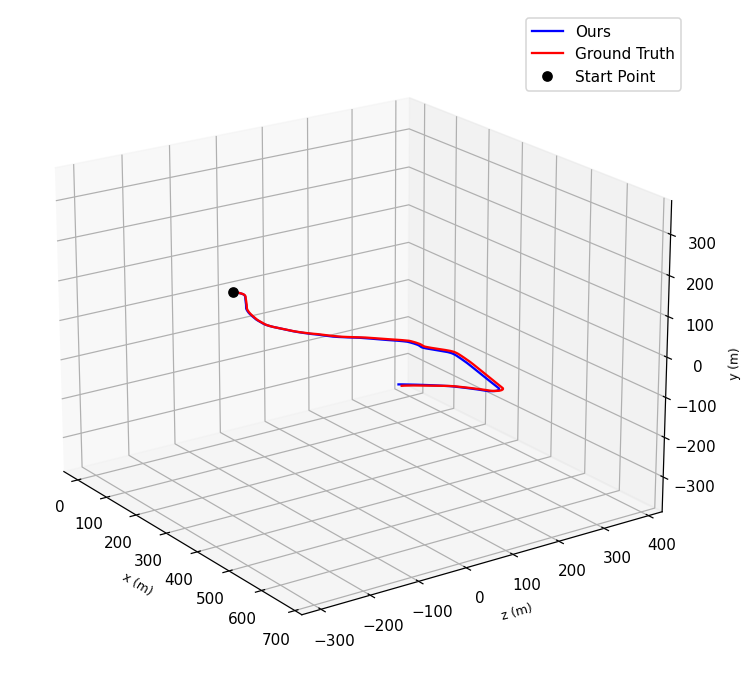}
        \caption{3D trajectory of seq.10}
    \end{subfigure}
    
    \caption{3D trajectory results on KITTI Seq. On the KITTI dataset, 07-10 shows the comparison between our DVLO4D with LOAM, Full LOAM, and Ground Truth.}
    \label{fig:trajectory}
    \vspace{-5mm}
\end{figure}

\section{Experiment}

\subsection{Dataset and Metrics}

\noindent
\textbf{The KITTI odometry dataset}~\cite{geiger2013vision} is a widely recognized benchmark for evaluating odometry and SLAM systems. 
For our experiments, we focus on the monocular left camera image in conjunction with the LiDAR sensor data.
Sequences 00-10, containing 23,201 scans, include ground truth trajectories, of which we utilize sequences 00-06 for training and sequences 07-10 for testing. 

\noindent
\textbf{The Argoverse Dataset}~\cite{argo} is a comprehensive dataset developed for advancing research in autonomous driving, offering LiDAR points and stereo imagery. It contains 113 sequences tailored for tracking tasks, divided into 65 sequences for training, 24 for validation, and 24 for testing.

\noindent
\textbf{Metrics.} To evaluate our method, we adhere to the evaluation protocols established by PWCLO~\cite{wang2021pwclo}, utilizing two primary metrics: (1) The average sequence translational RMSE (expressed as a percentage). (2) The average sequence rotational RMSE (measured in degrees per 100 meters).

\subsection{ Implementation Details}
\noindent
\textbf{Data Preprocessing.} We adopt sparse sampling and data augmentation strategy for LiDAR points from~\cite{wang2022efficient}. Given the significant spatial range difference between the camera and LiDAR, we designed a fusion mask to indicate which LiDAR queries can be fused with the image features. 

\noindent
\textbf{Parameters.} Experiments were performed on an NVIDIA RTX4090. We trained our model with 300 epochs and employed the Adam optimizer with $\beta_1 = 0.9$ and $\beta_2 = 0.999$. The initial learning rate was set to 0.001, with exponential decay every 13 epochs until it reached 0.00001. A batch size of 8 was used. The values for $\alpha_l$ in four layers were 1.6, 0.8, 0.4, and 0.8, and \( \beta \) was 0.8. The initial values of the learnable parameters $k_t$ and $k_q$ were 0.0 and -2.5, respectively. The number of LiDAR queries at each feature level was 116, 228, 904, and 3600, respectively. The KITTI training sequences (00-06) were divided into multiple video clips, each with a duration of \( T_C = 60 \) (6s) for the Temporal Interaction and Update module, and further segmented into sub-clips of \( T_s = 3 \) (0.3s) for collective average loss. The maximum history frame length \( T_h \) was set to 30 (3s).

\subsection{Results}
\noindent
\textbf{Comparison with Visual/LiDAR and Multi-Modal Odometry.} 
We evaluated our DVLO4D model against state-of-the-art visual odometry (VO), LiDAR odometry (LO) and learning-based multi-modal odometry methods on the KITTI dataset. As shown in Table~\ref{table:kitti_comparison}, our approach consistently improves performance across most sequences. Compared to VO methods like DFVO~\cite{zhan2021df} and Cho et al.~\cite{cho2023dynamic}, DVLO4D reduces the mean translation error \( t_{rel} \) and rotation error \( R_{rel} \) by more than 50\% on sequences 07-10. Against the state-of-the-art LiDAR odometry method EfficientLO~\cite{wang2022efficient}, DVLO4D reduces the mean translation error by 18\%. 

In comparison with learning-based multi-modal odometry methods like An et al.~\cite{an2022visual} H-VLO~\cite{aydemir2022h}, and methods in Table~\ref{tab:learning_based_comparison}, even though these methods are trained on broader datasets, DVLO4D surpasses them by reducing the mean translation error by 53\%. Additionally, compared to the state-of-the-art multi-modal odometry method DVLO~\cite{liu2024dvlo}, our model reduces the mean translation error by 12\% while maintaining a competitive rotation error of 0.37°/100m, showcasing DVLO4D's ability to leverage temporal information and multi-modal fusion effectively.

\noindent\textbf{Evaluation on the Argoverse dataset.} We evaluate the generalization ability of our method on Argoverse~\cite{argo}, following the protocol in~\cite{zhang2024dslo}, using Absolute Trajectory Error (ATE) and Relative Pose Error (RPE). As shown in Table~\ref{tab:argo-acc}, DVLO4D outperforms four geometry-based odometry methods~\cite{pylidar, ALOAM} without mapping, and surpasses the learning-based DSLO \cite{zhang2024dslo} by 20\% in ATE.


\noindent\textbf{Qualitative results}.
We present 3D trajectory visualizations based on our estimated poses in Fig~\ref{fig:trajectory}. The results demonstrate that our odometry method closely follows the ground truth trajectory. We also compared the 3D trajectory between the classical LOAM~\cite{zhang2014loam} and our method, as shown in seq.07. Despite being only the front end of a SLAM system without mapping, our method outperforms LOAM with mapping, demonstrating superior localization performance.

\noindent\textbf{Inference Speed}.
As shown in Table~\ref{table:inference_time}, we compare the inference speed of our DVLO4D against other multi-modal odometry methods. Efficiency plays a crucial role in real-time SLAM systems. Given that the LiDAR points in the KITTI dataset are recorded at a 10Hz frequency, most existing multi-modal methods~\cite{huang2020lidar,leutenegger2013keyframe,wang2021dv,shu2022multi,liu2024dvlo} struggle to meet the real-time application threshold of 100 ms. In contrast, our method achieves an inference time of only 82 ms, indicating its strong potential for real-time deployment.

\subsection{Handling Temporal Misalignment and Noise Robustness}

\noindent\textbf{Temporal Misalignment (10Hz vs 5Hz).} As shown in Table~\ref{tab:robustness}, when downsampling the LiDAR to 5Hz, our method exhibits a slight increase in mean error from \(0.73\) to \(0.75\) \((2.7\%)\), while EfficientLO shows a larger increase from \(0.86\) to \(0.99\) \((15.1\%)\). This indicates that our approach handles temporal misalignment with more robustness.

\noindent\textbf{Robustness to Noise.}  
With added noise, our method's mean error increases from \(0.73\) to \(0.75\) \((2.7\%)\), whereas EfficientLO's error rises from \(0.86\) to \(0.97\) \((12.8\%)\). This demonstrates that our method is more robust to sensor noise.

\subsection{Ablation Studies}
\begin{table}[t]
\centering
\caption{Ablation of multi-modal fusion strategies. }
\label{tab:local_fusion_ablation}
\scriptsize  
{\renewcommand{\arraystretch}{1.1}
\setlength{\tabcolsep}{3pt}
\resizebox{1\linewidth}{!}{%
\fontsize{8pt}{8pt}\selectfont
\begin{tabular}{c|cc|cc|cc|cc|cc|c}
\hline
\multirow{2}{*}{\textbf{Method}} & \multicolumn{2}{c|}{\textbf{07}} & \multicolumn{2}{c|}{\textbf{08}} & \multicolumn{2}{c|}{\textbf{09}} & \multicolumn{2}{c|}{\textbf{10}} & \multicolumn{2}{c|}{\textbf{Mean}} & \textbf{Speed} \\ \cline{2-11}
 & \textit{t\textsubscript{rel}} & \textit{r\textsubscript{rel}} & \textit{t\textsubscript{rel}} & \textit{r\textsubscript{rel}} & \textit{t\textsubscript{rel}} & \textit{r\textsubscript{rel}} & \textit{t\textsubscript{rel}} & \textit{r\textsubscript{rel}} & \textit{t\textsubscript{rel}} & \textit{r\textsubscript{rel}} & (ms) \\ \hline
Attention~\cite{liu2021swin} & 0.44 & 0.35 & 1.03 & 0.43 & 1.13 & 0.44 & 0.92 & 0.47 & 0.88 & 0.43 & \hfil180 \\ \hline
CNN~\cite{huang2020epnet} & 0.48 & 0.39 & 1.13 & 0.52 & 0.95 & 0.43 & 0.92 & 0.51 & 0.87 & 0.46 & \hfil85 \\ \hline
Clustering~\cite{liu2024dvlo} & 0.44 & 0.37 & 0.97 & 0.37 & 0.89 & 0.40 & 0.91 & 0.49 & 0.80 & 0.41 & \hfil94 \\ \hline
Query-based & \textbf{0.43} & \textbf{0.32} & \textbf{0.95} & \textbf{0.36} & \textbf{0.77} & \textbf{0.33} & \textbf{0.76} & \textbf{0.46} & \textbf{0.73} & \textbf{0.37} & \hfil\textbf{82} \\ \hline
\end{tabular}%
}}
\end{table}


\noindent
\textbf{Impact of Sparse Query Fusion (SQF).} Removal of the SQF module results in a significant performance drop. As shown in Table~\ref{table:ablation_tiu_sqf_cal}, the mean translation error (\(t_{rel}\)) increases by 34.2\% (from 0.73 to 0.98) and the rotation error (\(r_{rel}\)) rises by 54.1\% (from 0.37 to 0.57), proving that SQF is essential for effective multi-modal fusion by establishing fine-grained point-to-pixel correspondences. Moreover, Table~\ref{tab:local_fusion_ablation} shows that query-based fusion achieves the best accuracy and fastest inference speed (82 ms), outperforming convolution-based~\cite{huang2020epnet}, attention-based~\cite{liu2021swin}, and cluster-based~\cite{liu2024dvlo} methods. 

\noindent
\textbf{Effect of TIU.} 
When the TIU module is removed in Table~\ref{table:ablation_tiu_sqf_cal}, \( t_{rel} \) increases by 26\% (from 0.73 to 0.92) and \( r_{rel} \) by 45.9\% (from 0.37 to 0.54). The TIU plays a vital role in improving pose initialization and enhancing robustness by allowing the model to adjust its predictions. In addition, as shown in Table~\ref{tab:sub-clip-length}, increasing \( T_h \) from 15 to 30 leads to performance improvements however, continuing to increase \( T_h \) to 45 results in diminishing returns.

\noindent
\textbf{Contribution of CAL.} 
Without CAL, the results in Table~\ref{table:ablation_tiu_sqf_cal} shows the mean translation error (\(t_{rel}\)) increases by 13.7\% (from 0.73 to 0.83) and the rotation error (\(r_{rel}\)) rises by 32.4\% (from 0.37 to 0.49), demonstrating CAL's essence for improving temporal learning by aggregating losses over multiple frames. Also, increasing the sub-clip length \( T_s \) leads to improvements in performance in Table~\ref{tab:sub-clip-length}. 


\begin{table}[t]
\centering
\caption{The impact of various sub-clip length \( T_s \), and maximum history frame length \( T_h \). }
\label{tab:sub-clip-length}
\scriptsize
{\renewcommand{\arraystretch}{1.1}
\setlength{\tabcolsep}{5pt}
\resizebox{0.485\textwidth}{!}{%
\fontsize{8pt}{8pt}\selectfont
\begin{tabular}{c|c|cc|cc|cc|cc|cc}
\hline
 \multirow{2}{*}{\textbf{\( T_s \)}} & \multirow{2}{*}{\textbf{\( T_h \)}} & \multicolumn{2}{c|}{\textbf{07}} & \multicolumn{2}{c|}{\textbf{08}} & \multicolumn{2}{c|}{\textbf{09}} & \multicolumn{2}{c|}{\textbf{10}} & \multicolumn{2}{c}{\textbf{Mean}} \\ \cline{3-12}
  &  & \textit{t\textsubscript{rel}} & \textit{r\textsubscript{rel}} & \textit{t\textsubscript{rel}} & \textit{r\textsubscript{rel}} & \textit{t\textsubscript{rel}} & \textit{r\textsubscript{rel}} & \textit{t\textsubscript{rel}} & \textit{r\textsubscript{rel}} & \textit{t\textsubscript{rel}} & \textit{r\textsubscript{rel}} \\ \hline
1 & 30 & 0.47 & 0.39 & 1.03 & 0.41 & 0.98 & 0.52 & 0.83 & 0.64 & 0.83 & 0.49 \\ \hline
3 & 15 & \textbf{0.42} & 0.41 & 0.99 & 0.40 & 0.81 & \textbf{0.32} & 0.82 & 0.47 & 0.76 & 0.40 \\ \hline
3 & 30 & 0.43 & 0.32 & \textbf{0.95} & \textbf{0.36} & \textbf{0.77} & 0.33 & \textbf{0.76} & \textbf{0.46} & \textbf{0.73} & \textbf{0.37} \\ \hline
3 & 45 & 0.45 & \textbf{0.30} & 0.97 & 0.39 & 0.80 & 0.38 & 0.79 & 0.49 & 0.75 & 0.39 \\ \hline
\end{tabular}%
}}
\vspace{-5mm}
\end{table}

\section{Conclusion}
In this paper, we introduced DVLO4D, a novel visual-LiDAR odometry framework that leverages Sparse Query Fusion (SQF) and Temporal Interaction and Update (TIU) modules to improve odometry accuracy and robustness. Our method effectively fuses multi-modal data and leverages temporal information from past frames, leading to significant performance gains compared to state-of-the-art visual-LiDAR odometry methods. Through extensive experiments on the KITTI dataset, we demonstrated that DVLO4D achieves state-of-the-art results in both translation and rotation errors while maintaining real-time inference speed.

\vspace{6pt}
\noindent
\textbf{Acknowledgments}. This work was partially supported by the EU HORIZON-CL42023-HUMAN-01-CNECT XTREME (grant no. 101136006), and the Sectorplan Beta-II of the Netherlands.








\pagebreak
\bibliographystyle{IEEEtran}
\bibliography{bibliography}

\end{document}